\documentclass[runningheads]{llncs}
\usepackage[T1]{fontenc}
\usepackage{graphicx}
\usepackage[misc]{ifsym}
\usepackage{wrapfig}
\usepackage{orcidlink}
\usepackage{booktabs}
\usepackage{multirow}
\usepackage{soul}
\usepackage{xcolor}
\sethlcolor{yellow}

\begin{document}
%
\title{Optimization of sim-to-real transfer in the humanoid robot NICO}
\titlerunning{Optimization of sim-to-real transfer}
%
\author{Juraj Gavura
\and
Igor Farka\v{s}\textsuperscript{\orcidlink{0000-0003-3503-2080} \Letter} 
}
\authorrunning{J. Gavura and I. Farka\v{s}}
\institute{Department of Applied Informatics \\ Comenius University Bratislava, Slovakia 
\\ 
\email{gavura4@uniba.sk, farkas@fmph.uniba.sk} 
}
\maketitle  
\begin{abstract}
Robotic grasping requires accurate coordination between visual perception, object localization, inverse kinematics, and hand control. However, when movements planned in simulation are executed on a physical robot, the sim-to-real gap can cause small positioning errors that prevent successful grasping. In our previous work, we introduced a low-cost haptic calibration method that improved 2D reaching accuracy of the humanoid robot NICO. In this paper, we extend this approach from reaching to tabletop object grasping by adding YOLO-based object and hand detection, stereo vision-based localization using the robot's built-in low-resolution fisheye cameras, and task-specific corrections for grasp execution. Together, these components form a novel calibration-based grasping pipeline that does not require RGB-D cameras, motion capture, or external tracking systems. We also implemented a visual feedback model that aligns the robot hand with the detected object before grasping. Our results show that the fully nonlinear calibration model achieved the best performance inside the calibrated area, while the visual feedback model achieved the highest overall grasping success across the full tabletop workspace.

\keywords{Grasping \and Sim-to-real transfer \and Humanoid robot \and Stereo vision \and Visual feedback}
\end{abstract}
\section{Introduction}

Robotic grasping is one of the fundamental capabilities required for autonomous object manipulation. In humanoid robots, this task requires the integration of visual perception, object localization, inverse kinematics (IK), and hand control. Even when a robot can accurately reach a target point in simulation, the same movement may fail on the physical platform due to the sim-to-real gap caused by mechanical inaccuracies, actuator imperfections, sensor noise, or simplified assumptions in the robot model.

We addressed this problem in our previous work \cite{gavura2025} by introducing a low-cost calibration method based on haptic feedback from a touchscreen. The robot touched a set of target points on the screen, and the recorded contact positions were used to train correction models that mapped real-world target positions to suitable simulated coordinates. This approach improved the reaching accuracy of the Neuro-Inspired COmpanion (NICO) humanoid robot. However, reaching a point is only the first step toward successful manipulation, since grasping additionally requires object detection, localization, hand positioning, and reliable execution of the grasp.

In this paper, we extend this calibration-based sim-to-real transfer approach from 2D reaching to 2D object grasping. We propose a complete grasping pipeline that combines calibrated correction models,
object and hand detection based on You Only Look Once (YOLO) \cite{tian2025}, stereo vision-based localization, IK control, and a grasping procedure for tabletop objects. The vision module is built on low-resolution fisheye cameras, which makes the task challenging due to image distortion and limited visual detail. In addition, we design and implement a visual feedback model that uses the implemented perception and localization components to align the robot hand with the detected object before executing the grasp. Compared to approaches relying on  RGB with depth (RGB-D) cameras, motion capture systems, or complex external sensors, our method continues in the low-cost direction of our previous work and uses only the robot's built-in cameras together with calibration-based correction. The results show that the proposed pipeline improves grasping performance in the calibrated workspace, while visual feedback further increases robustness across the full tabletop area.

\section{Related work}

Vision-based robotic grasping usually combines object localization, pose or depth estimation, grasp planning, and robot control \cite{du2021}. Learning-based methods often predict grasp poses or grasp quality directly from visual input \cite{kleeberger2020}, but their performance can strongly depend on the perception setup, object appearance, background, illumination, and camera noise \cite{rameshbabu2025}. In our work, we focus on a simpler tabletop grasping scenario, where the target is localized from the robot cameras and executed using IK and calibrated correction models.

Many recent grasping methods rely on RGB-D cameras, point clouds, or large-scale datasets \cite{du2021}. GraspNet-1Billion provides a large RGB-D benchmark for general object grasping with a large number of annotated grasp poses \cite{fang2020}, while Dex-Net 4.0 uses synthetic point clouds and grasp robustness metrics to learn grasping policies \cite{mahler2019}. More recent work also uses foundation models to construct large-scale grasp datasets \cite{vuong2024}. These methods demonstrate the strength of depth-based and data-driven grasping, but they require dedicated sensors or large datasets. In contrast, we use only the built-in low-resolution fisheye stereo cameras of the humanoid robot.

Object detection is often used as the first step in robotic grasping systems. YOLO-based detectors have been combined with grasping networks for collaborative robots \cite{sun2024}, and YOLOv8 was used for target localization and grasping with the NAO humanoid robot using monocular ranging and error compensation \cite{jin2023}. Humanoid grasping has also been studied using visual servoing and stereo vision \cite{vahrenkamp2008}. These works are related to our vision module, where YOLO detects the target object and robot hand, while stereo vision estimates their positions in the robot workspace.

Several studies address sim-to-real transfer directly in robotic grasping. James et al. \cite{james2019} used randomized-to-canonical adaptation for data-efficient sim-to-real grasping, while Gäde et al. \cite{gade2022,gade2024} applied domain randomization and domain adaptation to visuomotor grasping with the NICO humanoid robot. Compared to these approaches, our work does not train an end-to-end grasping policy, but uses explicit localization, IK, and calibration-based correction.

Closed-loop visual feedback can improve grasp robustness by compensating for residual localization and execution errors. Morrison et al. \cite{morrison2018} continuously updated grasp predictions from depth images in real time, while Piacenza et al. \cite{piacenza2024} used visual feedback and adaptive sampling to refine grasp execution. Image-based visual servoing has also been used to control the final phase of closed-loop grasping \cite{haviland2020}. Inspired by these methods, we implement a simpler hand alignment model that aligns the detected hand with the detected object before grasping, using only the robot's stereo cameras and calibrated control pipeline.

\section{Materials and methods}

This work builds directly on our previous calibration approach \cite{gavura2025}. 
We use the same robot platform, simulation-control software, and calibrated correction models, which were adapted and further developed for the requirements of the tabletop grasping task.

\begin{wrapfigure}{r}{0.5\textwidth}
\vspace*{-7mm}
\includegraphics[width=0.5\textwidth]{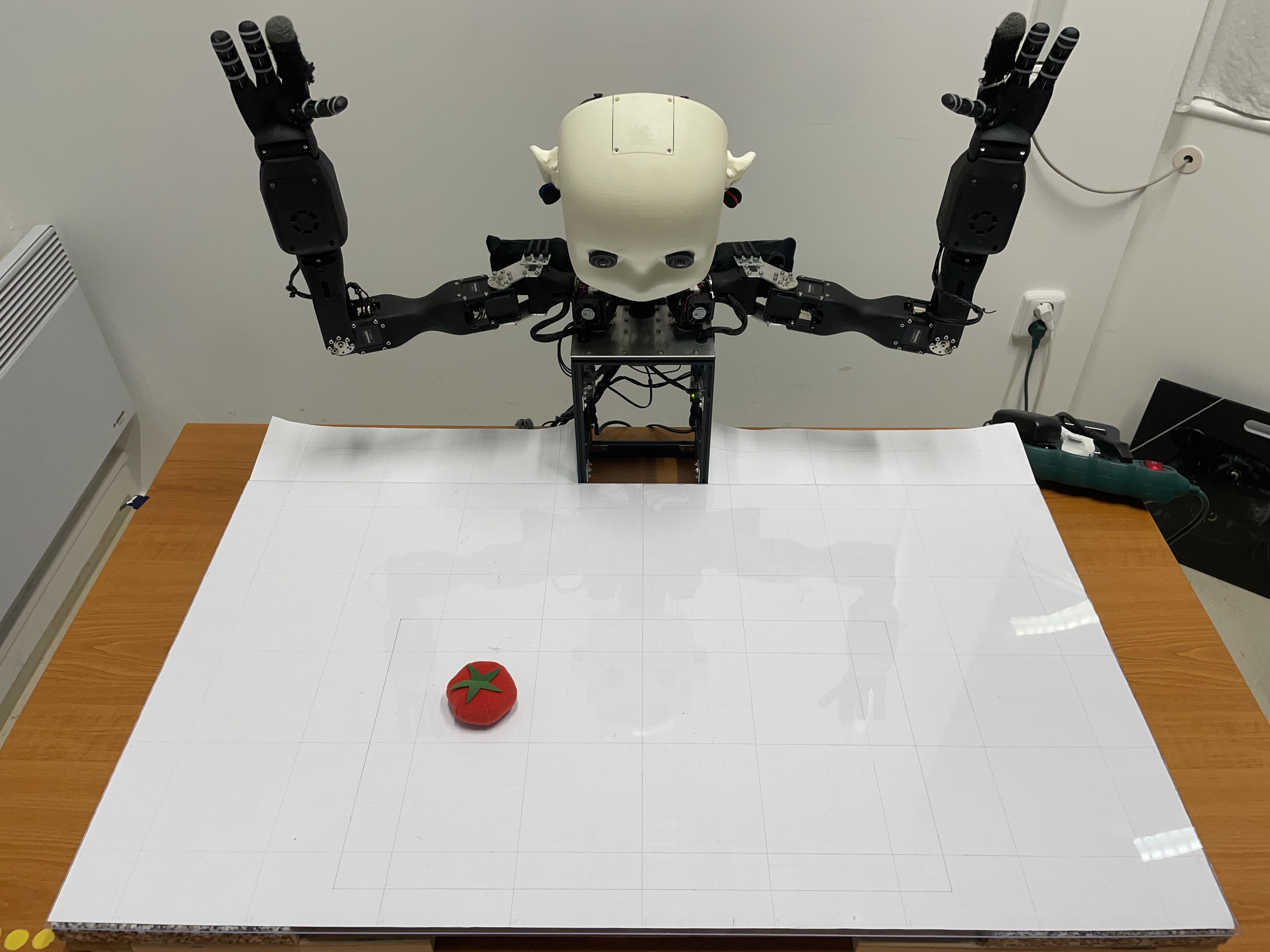}
\vspace*{-7mm}
\caption{NICO robot setup equipped with a paper grid used as ground truth for object placement in the grasping experiments.}
\vspace*{-7mm}
\label{nico_grid}
\end{wrapfigure}

\subsection{Robotic setup and software}

The experiments were performed with the humanoid robot NICO, using its right arm and four-finger child-sized hand.
A paper grid (Fig.~\ref{nico_grid}) provided a pseudo-ground truth for placing objects and evaluating grasping performance across the robot's reachable area.
For visual perception, we used the robot's built-in stereo camera setup, which provides low-resolution fisheye images of 640$\times$480 pixels.

The software framework was based on the NicoIK software\footnote{https://github.com/jgavura/NicoIK}, which links the simulated robot model with the real NICO robot and provides target positions for robot movements. The simulation module uses the PyBullet physics engine \cite{coumans2021} with a modified NICO unified robot description format (URDF) file and its inverse kinematics solver, while the real robot is controlled through the NicoMotion library \cite{nico_software}.

\subsection{Calibration models}

To reduce the discrepancy between target positions in the real workspace and corresponding positions used by the simulator and IK module, we used three calibration models. In the grasping task, these models transform the estimated object position on the table into a corrected target position for the robot hand.

The baseline model M1 uses a piecewise linear mapping based on the manually aligned simulation and interpolation of measured height values. M2 is a partially nonlinear neural-network correction model, where the horizontal target coordinates are predicted by a multilayer perceptron (MLP), while the height component is still obtained by interpolation. M3 is a fully nonlinear neural-network model that predicts the corrected three-dimensional target position directly. A detailed description of the calibration procedure, model training, and reaching evaluation is given in our previous work \cite{gavura2025}.

In this paper, the models are not retrained, but are used as correction modules within the grasping pipeline. This allows us to evaluate whether calibration models optimized for 2D reaching can also improve object grasping, where small residual positioning errors may have a larger effect on task success.

\subsection{Vision module}

Our vision module consists of two parts: object detection for identification and depth estimation to compute the object 3D position relative to the robot.

\subsubsection{Object and hand detection}

In addition to graspable objects, we also trained the detector to recognize the robot hands, which are later used for aligning the hand with the detected object before grasping.

We used the YOLO framework for its speed to accuracy balance. A custom dataset of 442 images was created and manually annotated into eight classes: six objects, left hand, right hand. An 80:20 ratio was used for the training to validation set split.
Several YOLO variants were finetuned on our custom dataset and evaluated on the validation set using mean average precision at intersection over union thresholds of 0.50--0.95 (mAP50--95) and inference time, as summarized in Table~\ref{yolo_table}. Although YOLO11m achieved the highest overall accuracy, YOLO12s was selected as a practical compromise between accuracy and real-time performance.

\begin{figure}[t]
\vspace*{-3mm}
\centering
\includegraphics[width=\textwidth]{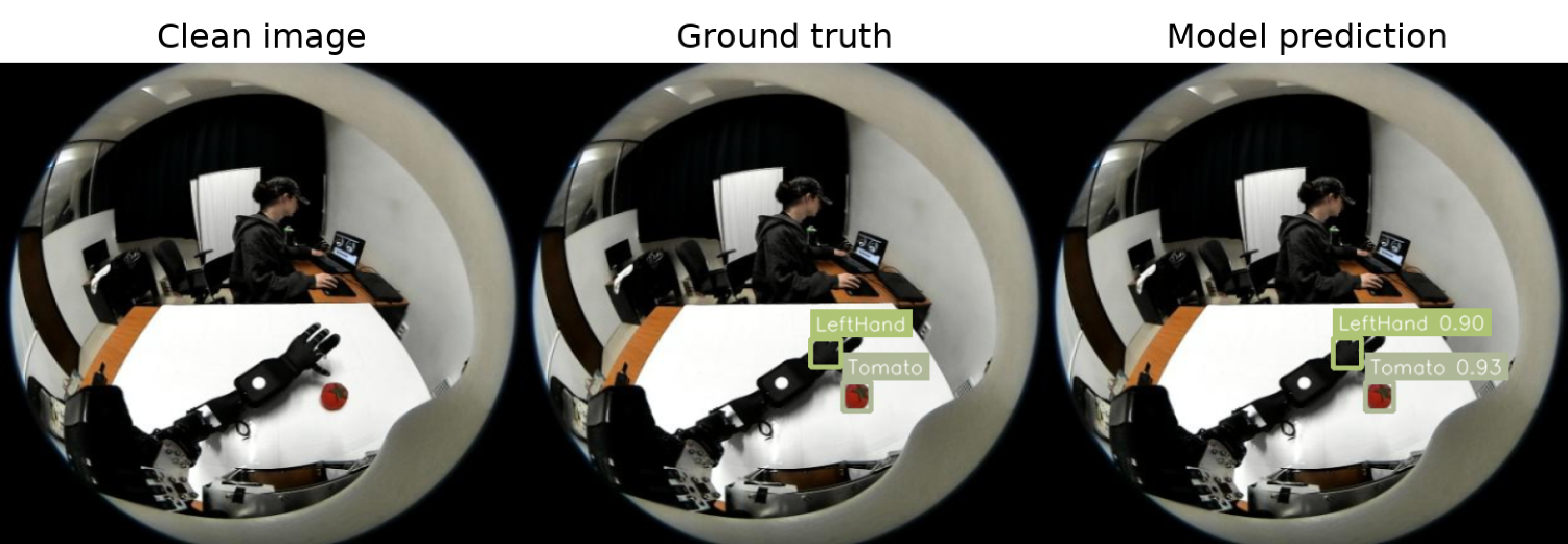}
\vspace*{-7mm}
\caption{Example of object detection results using the YOLO12s model. \textit{Left:} the original image; \textit{Middle:} ground truth annotations from the dataset; \textit{Right:} predictions by the trained model with confidence scores.}
\label{yolo_example_inference}
\end{figure}

The experiments were performed with a single object, a plush tomato, for its optimal size, high-contrast color and a regular shape, which simplifies grasp orientation.
Its detection together with the robot hand detection is shown in Fig.~\ref{yolo_example_inference}. Bounding boxes for hands were intentionally defined around the central part of the palm, so that its centroid better corresponded to the hand position used for alignment.

\begin{table}[t]
\centering
\caption{Comparison of YOLO models using mAP50--95 for Tomato class, Hand classes and all classes. Inference time is reported per image and does not include pre-processing and post-processing (n - nano, s - small, m - medium, SPP - Spatial Pyramid Pooling).}
\label{yolo_table}
\resizebox{0.7\textwidth}{!}{
\begin{tabular}{l|cccc|c}
\toprule
Model & Tomato & L-Hand & R-Hand & All & Inference [ms] \\
\midrule
YOLO12n & 0.884 & 0.526 & 0.587 & 0.772 & \textbf{6.20} \\
YOLO12s & 0.895 & 0.562 & 0.640 & 0.806 & 10.86 \\
YOLO12m & 0.902 & 0.592 & 0.594 & 0.813 & 25.63 \\
YOLO11n & 0.893 & 0.551 & 0.621 & 0.801 & 6.26 \\
YOLO11s & 0.893 & 0.568 & 0.623 & 0.805 & 10.09 \\
YOLO11m & \textbf{0.915} & \textbf{0.607} & 0.611 & \textbf{0.823} & 20.45 \\
YOLOv3 & 0.895 & 0.576 & \textbf{0.668} & 0.817 & 38.79 \\
YOLOv3-SPP & 0.901 & 0.537 & 0.643 & 0.804 & 39.30 \\
\bottomrule
\end{tabular}
}
\vspace*{-3mm}
\end{table}

\vspace{-5mm}
\subsubsection{Stereo vision-based localization}

\begin{wrapfigure}{r}{0.5\textwidth}
\includegraphics[width=0.5\textwidth]{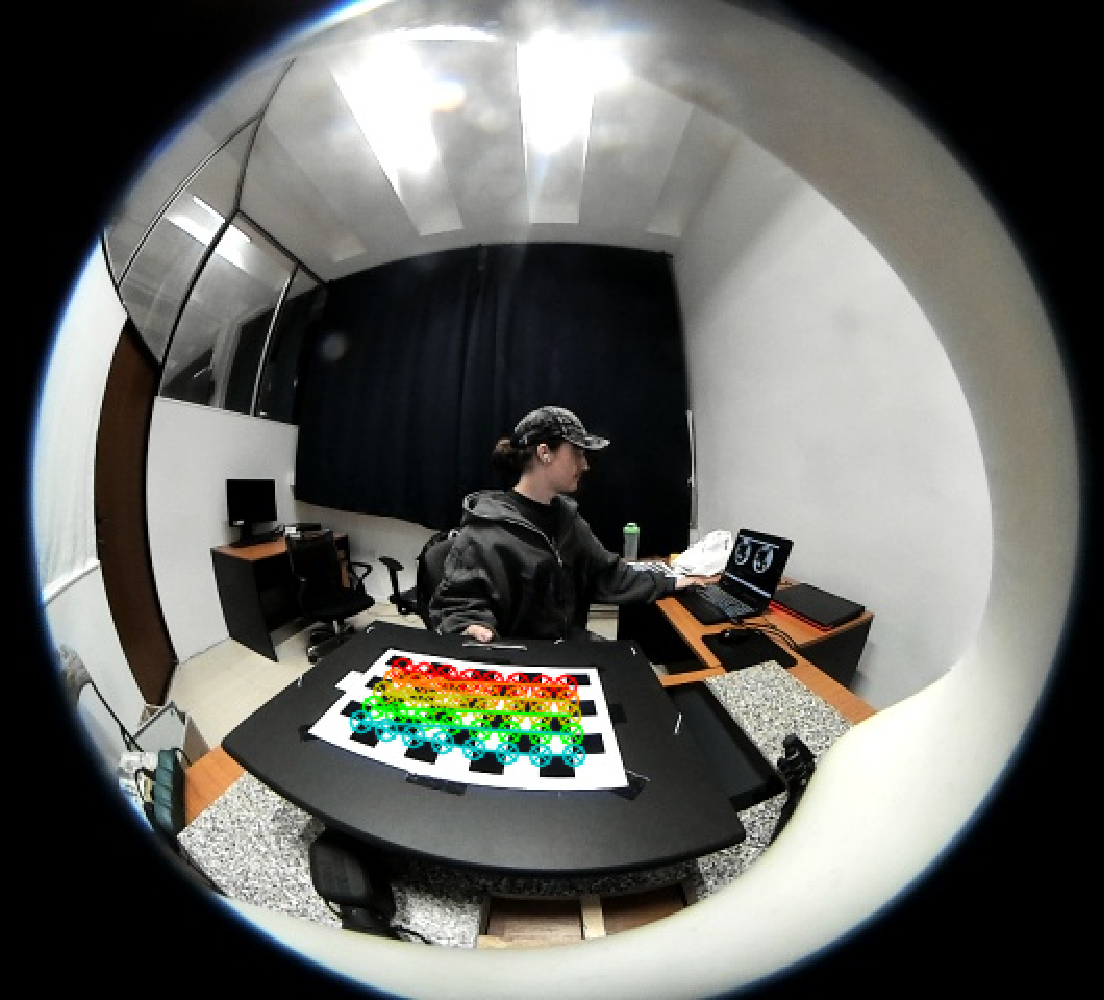}
\vspace*{-4mm}
\caption{Example of a stereo-vision calibration image with detected chessboard corners and their projected positions.}
\vspace*{-8mm}
\label{sv_corners_projection}
\end{wrapfigure}

After detecting the object in an image, stereo vision was used to estimate its position in the robot workspace. First, both cameras were calibrated using a chessboard pattern to obtain the intrinsic parameters and lens distortion coefficients. An example calibration image with detected and projected chessboard corners is shown in Fig.~\ref{sv_corners_projection}. The camera images were then undistorted and rectified, so that corresponding points appeared on the same horizontal image lines. From the rectified stereo pair, a disparity map was computed and used to estimate the depth of the detected object.

\begin{figure}[t]
\vspace*{-3mm}
\centering
\includegraphics[width=\textwidth]{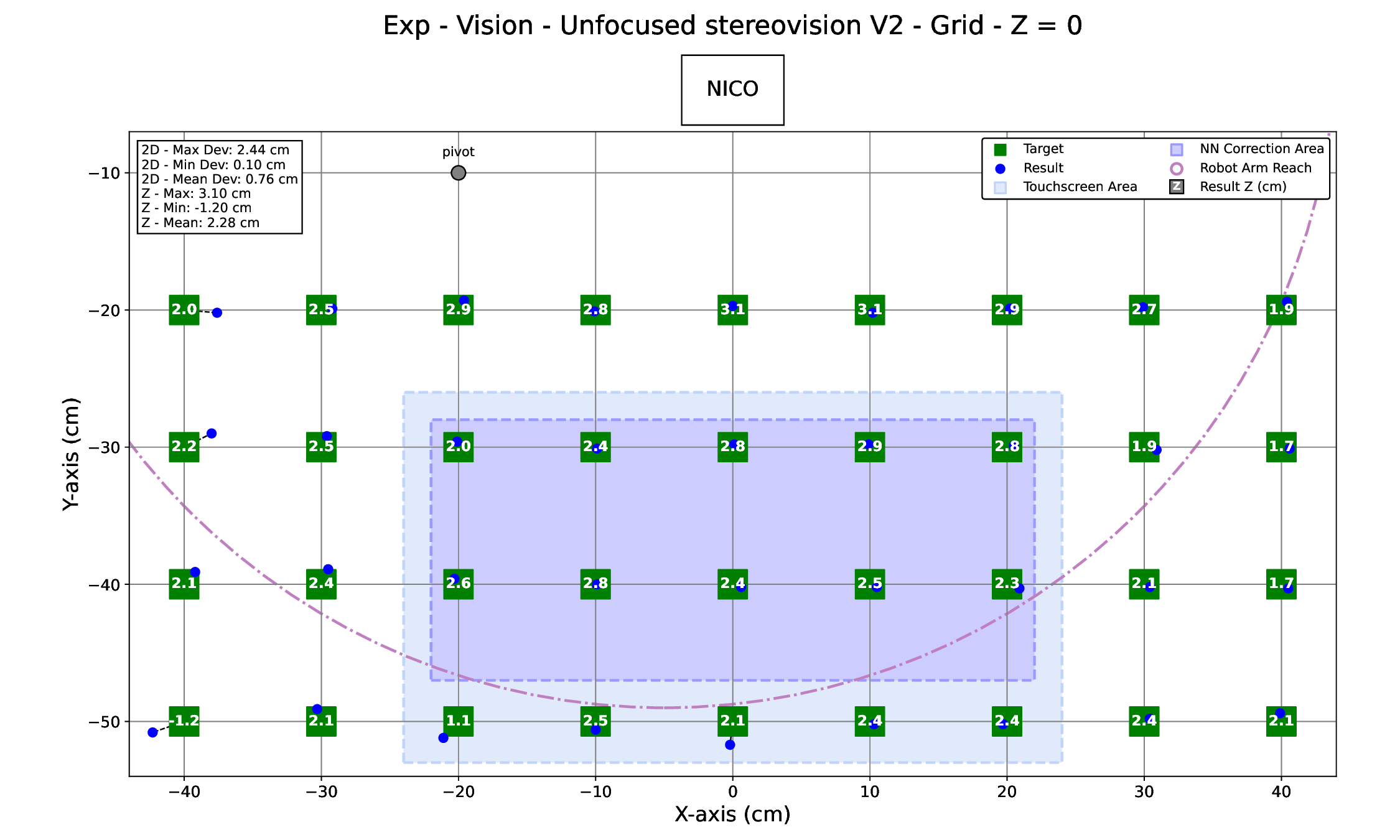}
\vspace*{-7mm}
\caption{Evaluation of the corrected stereovision localization method on a 9$\times$4 grid of target positions. The unfocused version uses the initial head pose, where NICO looks toward the center of the table. Green squares indicate target positions, blue points show the estimated positions, and the values inside the green squares represent the estimated $z$ coordinate in centimeters. The pivot point indicates the direction toward which the predictions are shifted during correction.} 
\label{vision_unfocused}
\vspace*{-5mm}
\end{figure}

The detected object center was first reconstructed in the camera coordinate frame and then transformed into the robot coordinate frame using the kinematic chain of the robot head.
During testing, we observed a localization drift likely caused by overheating cameras. While initially accurate, the error gradually increased and stabilized after 10 minutes, reaching about 6 cm for the farthest table positions.
Therefore, we applied an additional correction in the table plane by moving the predicted positions toward a fixed pivot point. The corrected localization results, evaluated on a 9$\times$4 grid of target positions distributed across the table, are shown together with the pivot location in Fig.~\ref{vision_unfocused}.

Using a fixed head position, where the robot kept its initial head pose during inference, the final stereo vision module achieved a mean 2D localization error 0.76 cm. Only a few outliers reached errors $>$$2$ cm, mostly at edge positions, many of which were outside the reachable area of the robot arm. We also tested an alternative version in which the robot moved its head toward the object before stereo inference, but the results were comparable. Therefore, we used the more practical fixed-head version, which also indicates that cameras were calibrated reliably even for objects outside the center of the image frame.

\subsection{Hand alignment using visual feedback}

In addition to the calibration models, we implemented a visual feedback method for aligning the robot hand with the detected object before grasping. First, the robot estimates the object position using the vision module and moves the hand to an approximate position above the object. Then, the vision module is used again to detect the robot hand and reconstruct its position using stereo vision.

Let $\mathbf{p}_{\rm obj}$ denote the estimated object position, $\mathbf{p}_{\rm hand}$ the estimated hand position, and $\mathbf{p}^{k-1}_{\rm IK}$ the previous target sent to the IK solver. In each iteration, the next IK target is shifted in the direction from the detected hand position toward the object:
\[
\mathbf{p}^{k}_{\rm IK} = \mathbf{p}^{k-1}_{\rm IK} + a_k(\mathbf{p}_{\rm obj} - \mathbf{p}_{\rm hand}),
\]
where the step size is gradually reduced as $a_k = 1/(2^{k-1})$.
This prevents large oscillations when the hand is already close to the object. The alignment is repeated until the horizontal distance between the detected hand position and the object position is below a predefined threshold (set to 1 cm). The hand is then lowered toward the object and the grasp is executed by closing the fingers.

To make hand detection more reliable, we also adjusted the hand pose during the alignment stage. The palm was rotated toward the camera so that the detector could observe the central part of the palm more consistently before estimating the hand position. 
This effectively prevented YOLO detection failures despite the lower general mAP for hands.

This method served as M4 in our comparison. In contrast to the M2 and M3 correction models, it did not use calibration-based target correction, but relied only on visual feedback to align the hand with the object before grasping.

\subsection{Grasping pipeline and experimental evaluation}

For successful grasp execution, we also applied two task-specific corrections: the vertical target correction and the palm yaw correction. Suitable z values and yaw angles were manually measured at an irregular set of reference points distributed across the robot workspace. For arbitrary target positions, both values were then estimated using radial basis function (RBF) interpolation. The roll and pitch angles of the palm were kept fixed to maintain a top-down grasping pose.

\begin{figure}[t]
\vspace*{-3mm}
\centering
\includegraphics[width=\textwidth]{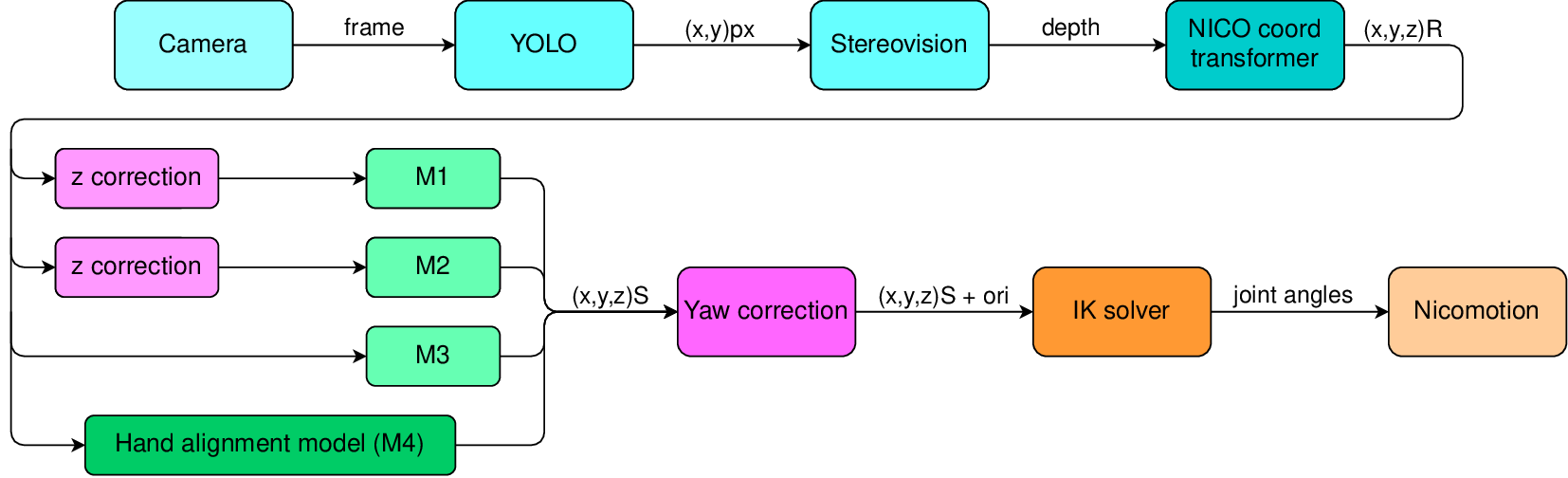}
\vspace*{-5mm}
\caption{Grasping pipeline of our approach.} 
\label{scheme}
\vspace*{-5mm}
\end{figure}

The complete grasping pipeline (Fig.~\ref{scheme}) starts with an image captured by the robot cameras. YOLO detector provides the object position in image coordinates, which is then used by stereo vision module to estimate depth. The detected point is transformed into robot's coordinate frame using the kinematic chain of the robot head. This real-world target position is then transformed into a simulated target position using one of the correction models. Finally, the yaw correction provides palm orientation, IK solver computes the required joint angles, and the movement is executed on the real robot through NicoMotion library.

We evaluated four grasping approaches using the models M1 to M4. The plush tomato was placed on 22 grid positions distributed across the reachable workspace of the robot arm. For each position and each model, three grasping attempts were performed, resulting in 66 attempts per model. In addition to the overall grasping success rate, we also evaluated the success rate inside the original neural-network correction area, which contained 10 target positions. This allowed us to compare both the performance in the calibrated region and the generalization of the models across the full tabletop workspace.

\section{Results}

\subsection{Calibration-based grasping}

\begin{figure}[t!]
\vspace*{-3mm}
\includegraphics[width=\textwidth]{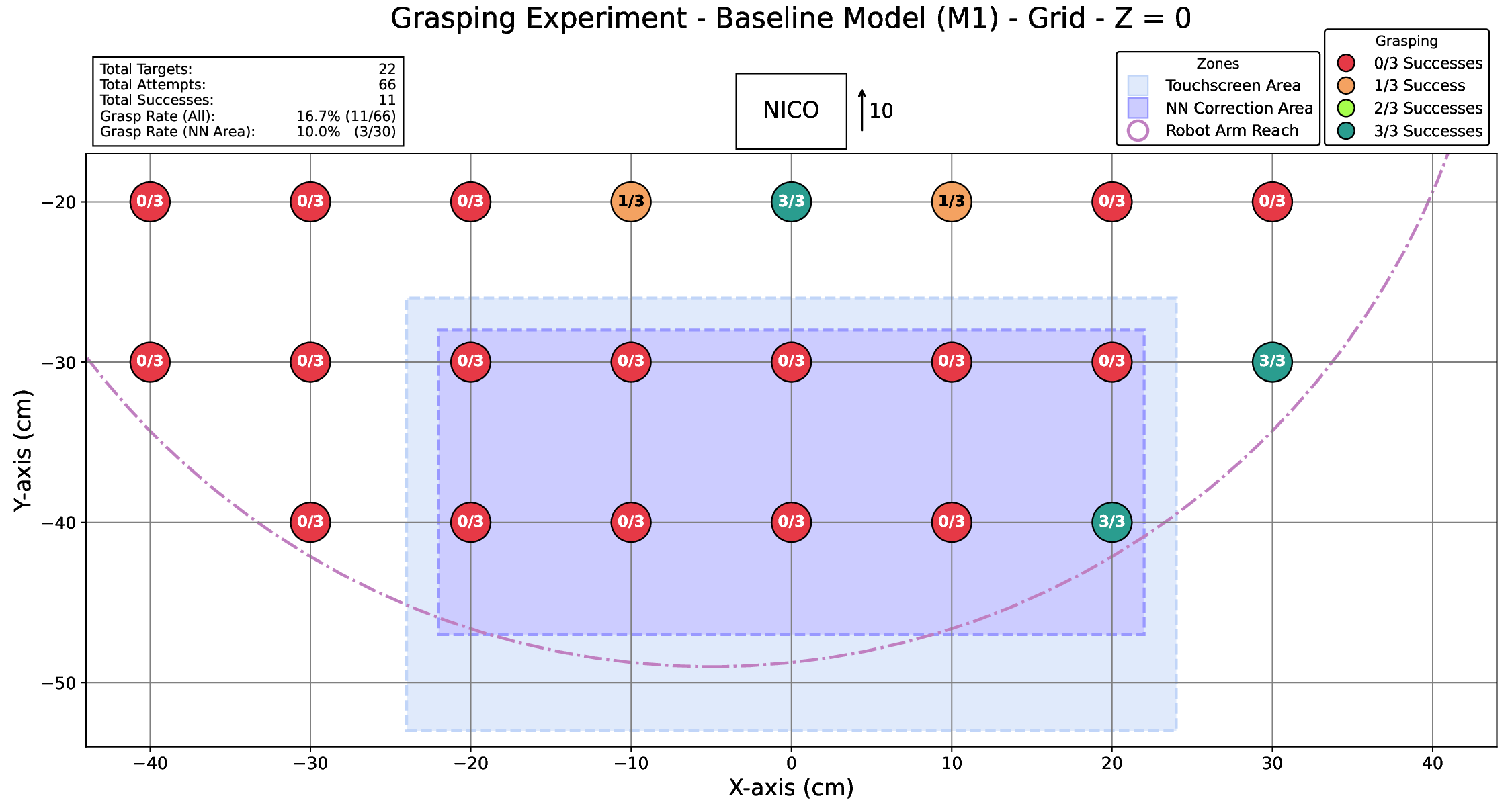}
\includegraphics[width=\textwidth]{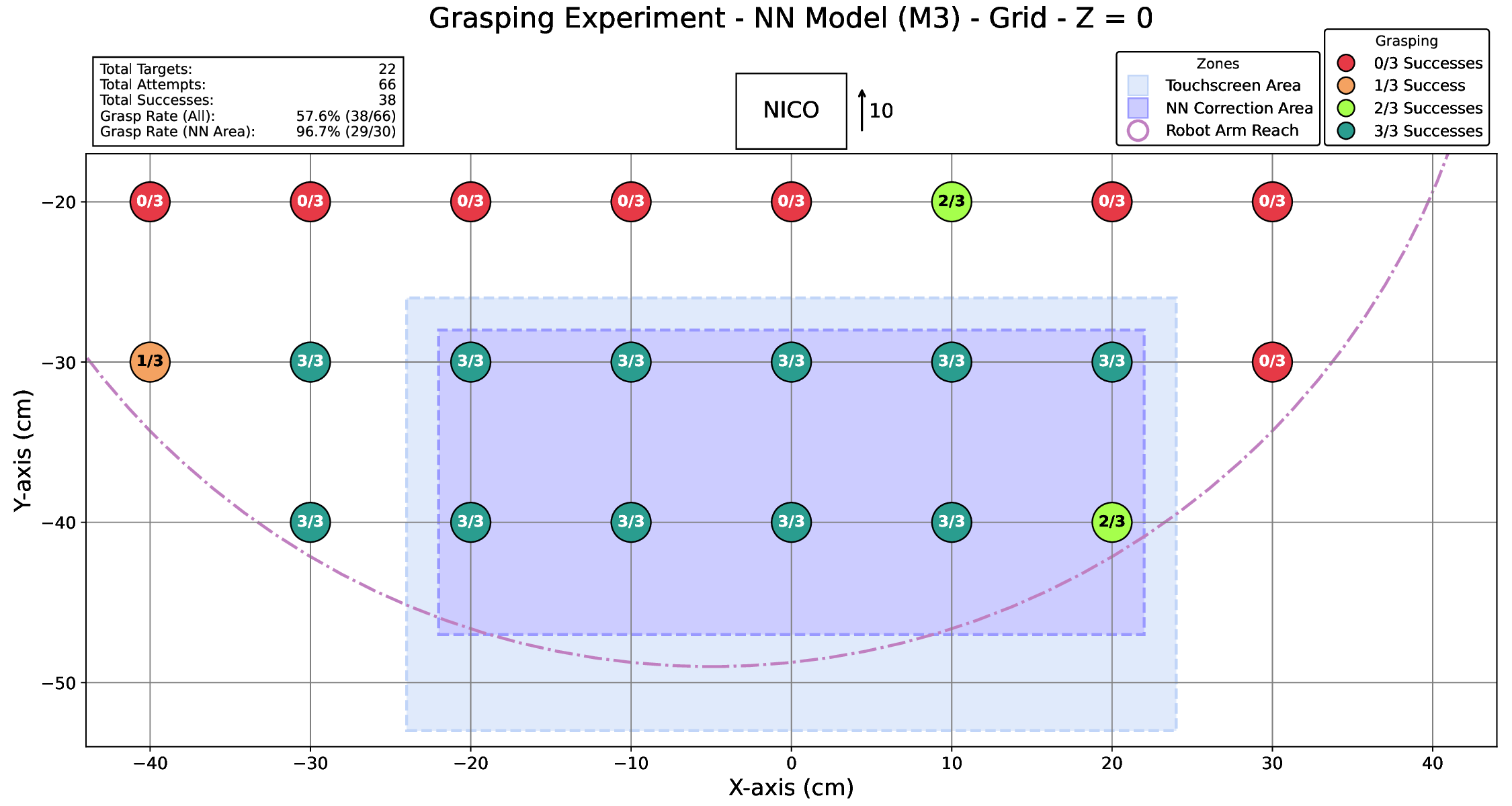}
\vspace*{-7mm}
\caption{Visualization of 2D grasping success rate for grid-based targets spread across the whole reachable area using the baseline model M1 (top) and NN correction model M3 (bottom). The plot shows the number of successful grasp attempts for each target position, together with the touchscreen area, NN correction area, and NICO arm reach. The box "NICO" denotes the robot position (to be shifted 10 cm up).} 
\label{grasp_baseline_vs_model3}
\vspace*{-5mm}
\end{figure}

We first evaluated the grasping performance of the calibration-based models to test whether the calibration also reduces the reality gap in the grasping task. The baseline model M1 is a purely linear model without additional correction in the horizontal $x$ and $y$ axes. As shown in the upper half of Fig.~\ref{grasp_baseline_vs_model3}, M1 achieved only 16.7\% success rate across the whole tested workspace, corresponding to 11 successful grasps out of 66 attempts. Inside the neural-network correction area, the success rate was 10.0\%, with 3 successful grasps out of 30 attempts. Although the hand often moved close to the object, the remaining positional error was usually large enough to prevent a stable grasp.

The best calibration-based performance was obtained with the fully nonlinear correction model M3, which uses a neural network to predict the complete corrected 3D target position, including the $z$ coordinate. As shown in the lower half of Fig.~\ref{grasp_baseline_vs_model3}, the main improvement was observed inside the neural-network correction area, where M3 achieved a 96.7\% success rate, corresponding to 29 successful grasps out of 30 attempts. This is the region for which training data were available and where the model was expected to be most effective. Across the whole tested workspace, M3 achieved a 57.6\% success rate, with 38 successful grasps out of 66 attempts. However, the results around the original touchscreen area did not improve substantially, which suggests that the neural network did not extrapolate reliably outside the region covered by the training data.

\begin{figure}[t!]
\vspace*{-5mm}
\includegraphics[width=\textwidth]{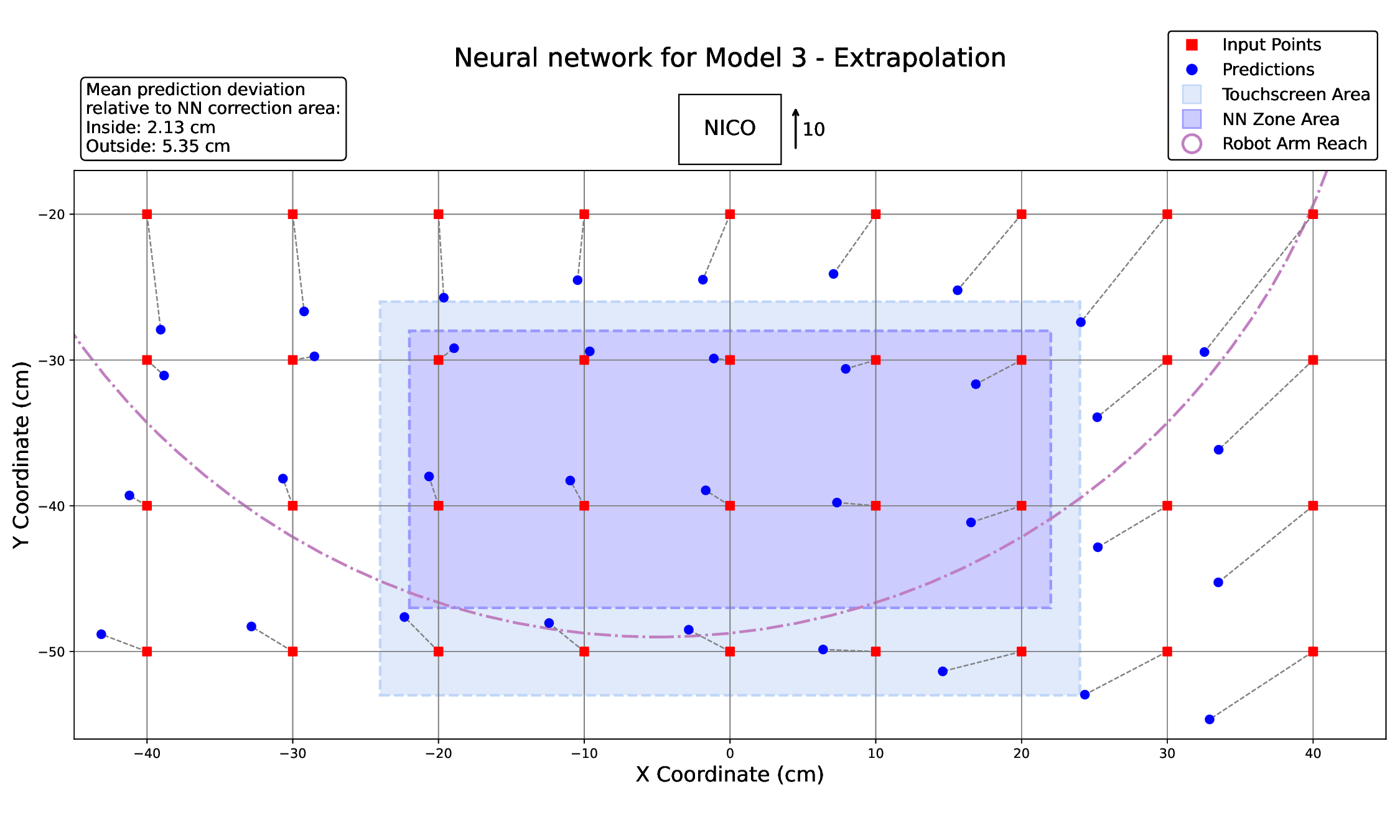}
\vspace*{-10mm}
\caption{Visualization of predictions produced by M3 for input points distributed across the whole table. Red points represent input target positions and blue points show the predicted positions. The darker rectangle marks the region covered by training data used for the NN, whereas points outside this region require extrapolation. The box "NICO" denotes the robot position.} 
\label{model3_extrapolation}
\end{figure}

To verify this hypothesis, we visualized the predictions produced by the neural network in M3, including points outside the neural network (NN) correction area. As shown in Fig.~\ref{model3_extrapolation}, the mean displacement between input target positions and predicted simulated positions was only 2.14 cm inside the correction area, but increased to 5.45 cm outside it. The predictions therefore confirm that M3 extrapolated poorly in a large part of the tabletop workspace, usually overshooting the required correction. The only exception was the region to the left of the original touchscreen area, where the predictions followed the same trend as inside the calibrated region, which also explains why several grasping attempts in this part of the workspace were successful.

\begin{figure}[h!]
\includegraphics[width=\textwidth]{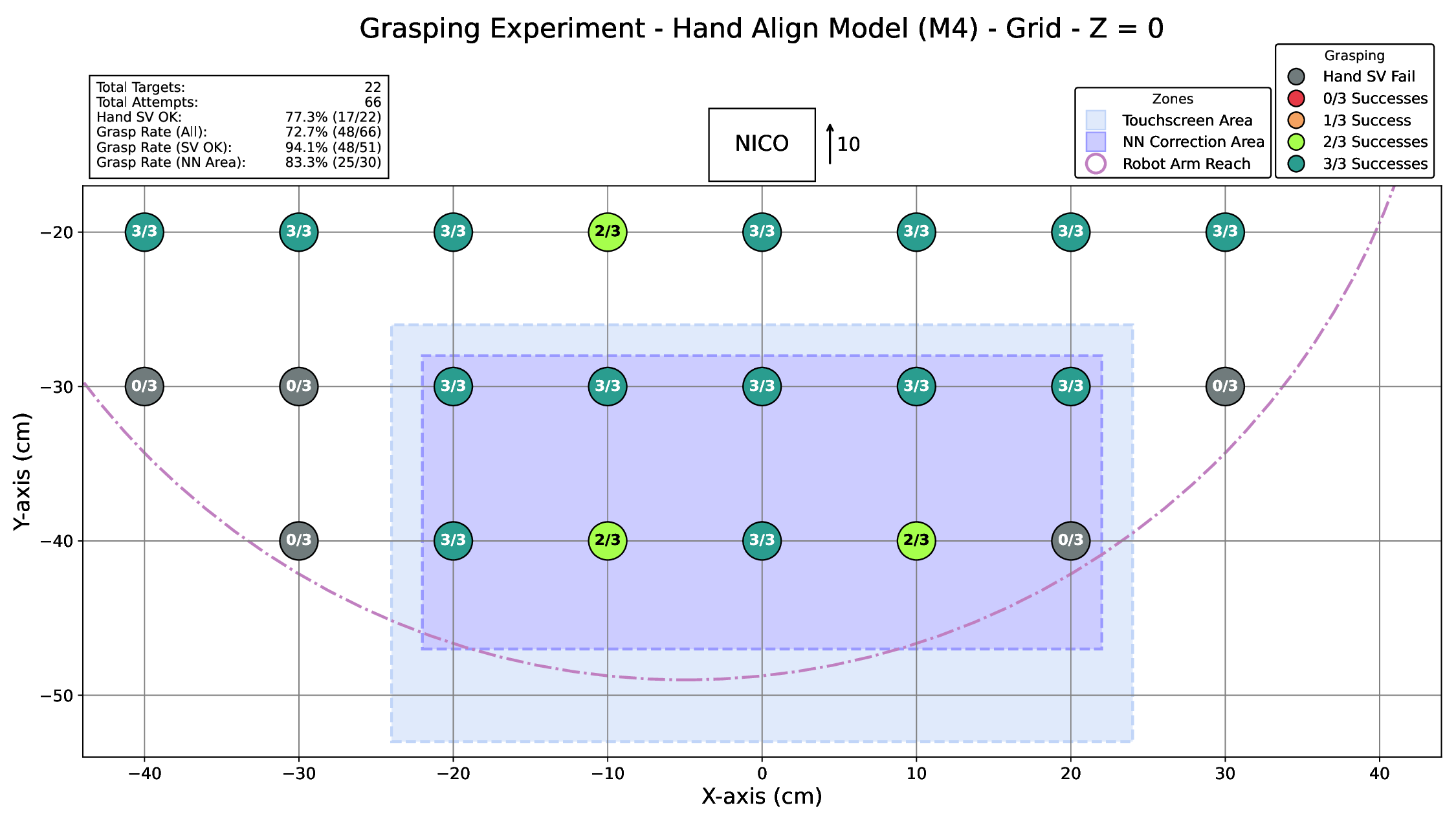}
\vspace*{-7mm}
\caption{Visualization of 2D grasping success rate for grid-based targets spread across the whole reachable area using the visual feedback hand alignment model. The plot shows the number of successful grasp attempts for each target position. Gray targets indicate positions where stereo vision substantially overestimated the hand position.} 
\label{grasp_align}
\vspace*{-7mm}
\end{figure}

\vspace{-2mm}
\subsection{Visual feedback and overall comparison}
\vspace{-1mm}

The visual feedback hand alignment model was evaluated as an alternative to the calibration-based correction models. Unlike M2 and M3, it did not use the calibrated target correction, but relied only on visual feedback to align the detected hand position with the detected object position before grasping. As shown in Fig.~\ref{grasp_align}, this approach improved grasping performance across the full tabletop workspace. Successful grasps were no longer concentrated only inside the neural-network correction area, but also appeared in regions outside the original touchscreen area. Overall, this model achieved a 72.7\% success rate, corresponding to 48 successful grasps out of 66 attempts.

However, the performance of this model was limited by the reliability of stereo vision for hand localization. Stereo vision correctly localized the hand in only 17 out of 22 target positions. In the remaining positions, stereo vision strongly overestimated the hand position, especially when there was no nearby background behind the hand and the visible background was one or two meters farther away. In such cases, the reconstructed hand position could not be used reliably for visual feedback. These positions are marked in Fig.~\ref{grasp_align} by gray color. When considering only the attempts where stereo vision localized the hand correctly, the model achieved a 94.1\% success rate, with 48 successful grasps out of 51 attempts. This suggests that the visual feedback approach could achieve even higher overall performance with more robust hand localization.

\begin{table}[t!]
\vspace*{-1mm}
\caption{Summary of grasping experiment statistics across all tested models.}
\vspace*{-3mm}
\label{models_table}
\resizebox{\textwidth}{!}{
\begin{tabular}{l|cccc}
\toprule
Statistic & Baseline M1 & M2 & M3 & Visual feedback \\
\midrule
Hand SV OK & -- & -- & -- & 77.3\% (17/22) \\
Grasp Rate (SV OK) & -- & -- & -- & 94.1\% (48/51) \\
Grasp Rate (NN Area) & 10.0\% (3/30) & 93.3\% (28/30) & \textbf{96.7\% (29/30)} & 83.3\% (25/30) \\
Grasp Rate (All) & 16.7\% (11/66) & 48.5\% (32/66) & 57.6\% (38/66) & \textbf{72.7\% (48/66)} \\
\bottomrule
\end{tabular}
}
\vspace*{-5mm}
\end{table}

The final comparison of all models is in Table~\ref{models_table}. The M2 model achieved lower performance than M3, which is consistent with the reaching results reported in our previous work. Inside the neural-network correction area, the best result was obtained by M3 with 96.7\% success rate. Across the whole tested workspace, the best result was obtained by the visual feedback model with 72.7\% success rate.

\vspace{-4mm}
\section{Discussion}
\vspace{-2mm}

Direct comparison with other robotic grasping studies is difficult, since platforms, objects, sensors, and evaluation protocols differ substantially. The most comparable work is  \cite{gade2024}, where sim-to-real grasping on NICO robot was also evaluated, but with side grasping of a cylindrical object using the left hand. Their best model achieved an 80.3\% success rate using domain adaptation and inherent IK.
Our visual feedback model achieved 72.7\% overall, while the fully nonlinear calibration model M3 reached 96.7\% inside the calibrated NN correction area.

Other sim-to-real and closed-loop grasping approaches also report strong results, but often rely on more complex learning-based systems. In \cite{james2019} 70\% zero-shot grasp success was achieved using randomized-to-canonical adaptation networks, increasing to 91\% after fine-tuning with 5,000 real-world grasps. In \cite{morrison2018} closed-loop generative grasping convolutional neural network (GG-CNN) with depth images was used and achieved 83\% success on unseen adversarial objects and 88\% on moving household objects. Our method also uses visual feedback, but instead of predicting grasp poses from depth images, it aligns the detected hand with the object using stereo cameras and a calibrated control pipeline.

The main limitation of the visual feedback model was stereo vision-based hand localization. In several positions, the hand position was strongly overestimated when the background behind the hand was too far from the robot. This reduced the overall success rate, although performance remained high when hand localization was correct. Improving hand segmentation or depth estimation would likely further improve the model.

In summary, we implemented a low-cost grasping pipeline combining object detection, stereo vision localization, and calibrated IK control without RGB-D cameras, motion capture, or external tracking. Although task-specific corrections were used to compensate for mechanical inaccuracies, the underlying methodology remains generalizable.
The results show that haptic calibration developed for 2D reaching can also improve object grasping, while visual feedback provides an effective alternative when calibration-based correction is not used.

\begin{credits}
 \subsubsection{\ackname}  We thank anonymous reviewers for detailed feedback. I.F. was supported by Slovak Research and Development Agency, project APVV-21-0105.
\end{credits}
%
%
%
\bibliographystyle{splncs04}
\bibliography{references}
%

\end{document}